# Real-Time Evaluation Models for RAG: Who Detects Hallucinations Best?

Ashish Sardana

Department of Computer Engineering
San Jose State University
San Jose, CA 95192
ashish.sardana@sjsu.edu

## Abstract

This article surveys Evaluation models to automatically detect hallucinations in Retrieval-Augmented Generation (RAG), and presents a comprehensive benchmark of their performance across six RAG applications. Methods included in our study include: LLM-as-a-Judge, Prometheus, Lynx, the Hughes Hallucination Evaluation Model (HHEM), and the Trustworthy Language Model (TLM). These approaches are all reference-free, requiring no ground-truth answers/labels to catch incorrect LLM responses. Our study reveals that, across diverse RAG applications, some of these approaches consistently detect incorrect RAG responses with high precision/recall. Code to reproduce our study is here: https://github.com/cleanlab/cleanlab-tools/tree/main/benchmarking_hallucination_model

## 1 Introduction

*Retrieval-Augmented Generation* enables AI to rely on company-specific knowledge when answering user requests [1]. While RAG reduces LLM hallucinations, they still remain a critical concern that limits trust in RAG-based AI. Unlike traditional Search systems, RAG systems occasionally generate misleading/incorrect answers. This unreliability poses risks for organizations that deploy RAG externally, and limits how much internal RAG applications get used.

Real-time *Evaluation Models* offer a solution, by providing a confidence score for every RAG response. Recall that for every user query: a RAG system retrieves relevant context from its knowledge base and then feeds this context along with the query into a LLM that generates a response for the user. Evaluation models take in the **response**, **query**, and **context** (or equivalently, the actual LLM prompt used to generate the response), and then output a score between 0 and 1 indicating how confident we can be that the response is correct. The challenge for Evaluation models is to provide *reference-free* evaluation (i.e. with no ground-truth answers/labels available) that runs in *real-time* [2].

Note that while *hallucination* sometimes refers to specific types of LLM errors, throughout we use this term synonymously with *incorrect* responses (i.e. what ultimately matters to users of a real RAG system).

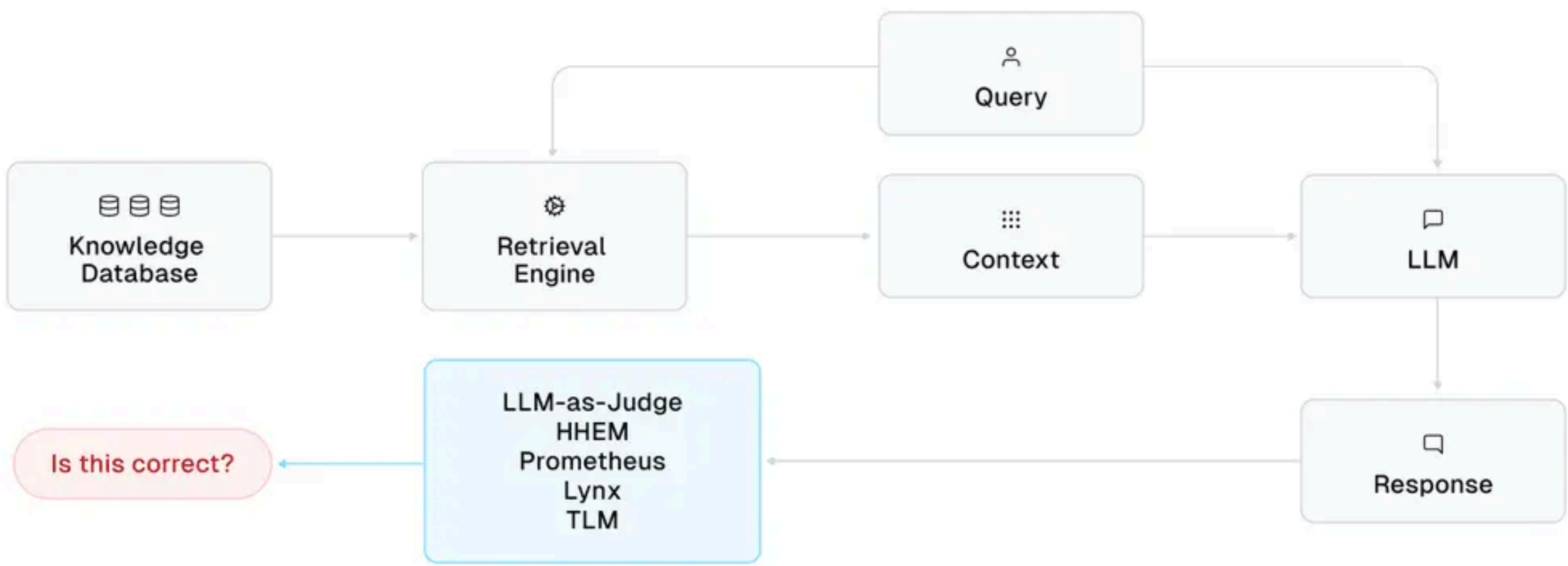

Figure 1. How Evaluation models are incorporated into a RAG system to automatically detect incorrect/hallucinated response

## 2 Popular Real-Time Evaluation Models

### 2.1 LLM as a Judge

LLM-as-a-judge (also called Self-Evaluation) [3] is a straightforward approach, in which the LLM is directly asked to evaluate the correctness/confidence of the response. Often, the same LLM model is used as the model that generated the response. One might run LLM-as-a-judge using a Likert-scale scoring prompt like this:

```
Context: {context}
Question: {question}
Answer: {response}
Evaluate how confident you are that the given Answer is a good and accurate
response to the Question.
Please assign a Score using the following 5-point scale:
1: You are not confident that the Answer addresses the Question at all, the
Answer may be entirely off-topic or irrelevant to the Question.
2: You have low confidence that the Answer addresses the Question, there are
doubts and uncertainties about the accuracy of the Answer.
3: You have moderate confidence that the Answer addresses the Question, the
Answer seems reasonably accurate and on-topic, but with room for
improvement.
4: You have high confidence that the Answer addresses the Question, the
Answer provides accurate information that addresses most of the Question.
5: You are extremely confident that the Answer addresses the Question, the
Answer is highly accurate, relevant, and effectively addresses the Question
in its entirety.
The output should strictly use the following template: Explanation: [provide
a brief reasoning you used to derive the rating Score] and then write
'Score: <rating>' on the last line.
```

The main issue with LLM-as-a-judge is that hallucination stems from the unreliability of LLMs, so directly relying on the LLM once again might not close the reliability gap as much as we'd like. That said, simple approaches can be surprisingly effective and LLM capabilities are ever-increasing.

### 2.2 Hughes Hallucination Evaluation Model (HHEM)

Vectara's Hughes Hallucination Evaluation Model [4] focuses on factual consistency between the AI response and retrieved context. Here we specifically consider HHEM version 2.1 [5]. This model's scores have a probabilistic interpretation, where a score of 0.8 indicates an 80% probability that the response is factually consistent with the context. For example:

```
context = 'I am in the United States'
response = 'I am in California'
hhem_score = 0.3  # factually inconsistent
```

These scores are produced by a Transformer model that Vectara trained to distinguish between hallucinated vs. correct responses from various LLMs over various context/response data. More specifically, this model processes (premise, hypothesis) inputs and outputs a score. For RAG, one can format these inputs like this:

```
premise = context + question
hypothesis = response
```

In our study, including the question within the premise improved HHEM results over just using the context alone.

## 2.3 Prometheus

Prometheus (specifically the recent Prometheus 2) [6] is a fine-tuned LLM model that was trained on direct assessment annotations of responses from various LLMs over various context/response data. This model is available in various sizes, such as 7B and 8 x 7B (mixture of experts), which are fine-tunes of Mistral [7] models. Here we focus on the *Prometheus2 8x7B* [8] (mixture of experts) model, as the authors reported that it performs better in direct assessment tasks. One can think of Prometheus as an LLM-as-a-judge that has been fine-tuned to align with human ratings of LLM responses.

Here is the specific prompt template used by Prometheus for direct assessment evaluation of LLM responses:

```
###Task Description:

An instruction (including passage and a question), a response to evaluate, and a score rubric representing an evaluation criteria are given.

1. Write a detailed feedback that assesses the response strictly based on the given score rubric, not evaluating in general.

2. After writing a feedback, write a score that is an integer between 1 and 5. You should refer to the score rubric.

3. The output format should look as follows: "(write a feedback for criteria) [RESULT] (an integer number between 1 and 5)"

4. Please do not generate any other opening, closing, and explanations.

###The instruction to evaluate:
{instruction}

 ###Response to evaluate:
{response}

###Score Rubrics:
{rubric}

###Feedback:
```

The instruction in the above prompt template expects both the context and the question which is prepared by concatenating strings:

```
instruction = "Passage:\n" + {passage} + "\nQuestion:\n" + {question}
```

In addition to a prompt template, Prometheus also requires a scoring rubric. This model is trained to work best with a 5-point Likert scale. In our study, we used the original scoring rubric format recommended by the Prometheus developers, and minimally modified the descriptions of each rating to suit each RAG use-case.

Here's an example of the scoring rubric used for the PubmedQA [9] dataset:

```
How accurately does the response reflect the evidence presented in the passage, maintaining appropriate scientific precision and avoiding unsupported claims or speculation?

Score 1: The response makes claims that directly contradict the passage, fabricates unsupported information, misuses technical terminology, introduces speculative mechanisms or implications, makes absolute claims without appropriate uncertainty, or drastically mismatches the passage's level of detail. The response fails to maintain scientific integrity and cannot be considered reliable.

Score 2: The response contains significant misinterpretations of evidence, unsupported extrapolations beyond data scope, imprecise use of technical terminology, inclusion of speculative details, overgeneralization of findings, or substantial deviation from the passage's level of detail. While some aspects may be accurate, the response's reliability is compromised by these issues.

Score 3: The response generally aligns with the main findings but includes minor unsupported details, slight misinterpretations, occasional imprecise terminology, reasonable but unsupported elaborations, missing some
```

```
limitations, or inconsistent detail level. While generally reliable, the response requires some scrutiny for complete accuracy.

Score 4: The response accurately reflects the evidence with only minor issues such as subtle extrapolations (though reasonable), rare imprecisions in technical terminology, occasional missing caveats, or slight variations in detail level. The response maintains good scientific integrity and can be considered largely reliable.

Score 5: The response perfectly reflects the presented evidence, maintains appropriate scientific uncertainty, uses precise technical terminology, avoids unsupported speculation, properly acknowledges limitations, and matches the passage's level of detail. The response maintains complete scientific integrity and can be fully relied upon as an accurate reflection of the passage.
```

## 2.4 Patronus Lynx

Like the Prometheus model, Lynx [9] is a LLM fine-tuned by Patronus AI to generate a PASS/FAIL score for LLM responses, trained on datasets with annotated responses. Lynx utilizes chain-of-thought to enable better reasoning by the LLM when evaluating a response. This model is available in 2 sizes, 8B and 70B, both of which are fine-tuned from the Llama 3 Instruct model [10]. Here we focus on the Patronus-Lynx-70B-Instruct model [11] because it outperforms the smaller 8B variant in our experiments.

Here is the specific prompt template used by Lynx:

```
Given the following QUESTION, DOCUMENT and ANSWER you must analyze the provided answer and determine whether it is faithful to the contents of the DOCUMENT. The ANSWER must not offer new information beyond the context provided in the DOCUMENT. The ANSWER also must not contradict information provided in the DOCUMENT. Output your final verdict by strictly following this format: "PASS" if the answer is faithful to the DOCUMENT and "FAIL" if the answer is not faithful to the DOCUMENT. Show your reasoning.

- QUESTION (THIS DOES NOT COUNT AS BACKGROUND INFORMATION):
{question}

- DOCUMENT:
{context}

- ANSWER:
{answer}

- Your output should be in JSON FORMAT with the keys "REASONING" and "SCORE":
{{"REASONING": <your reasoning as bullet points>, "SCORE": <your final score>}}
```

This model was trained on various RAG datasets with correctness annotations for responses from various LLMs. The training datasets included CovidQA, PubmedQA, DROP, FinanceBench, (all released in [9]) which are also present in our benchmark study. In our benchmark, **we omit Lynx results for these datasets that it already saw during its training**.

## 2.5 Trustworthy Language Model (TLM)

Unlike HHEM, Prometheus, and Lynx: Cleanlab's Trustworthy Language Model [12] (TLM) does not involve a custom-trained model. The TLM system is more similar to LLM-as-a-judge in that it can utilize any LLM model, including the latest frontier LLMs as soon as they are released. TLM is a wrapper framework on top of any base LLM that uses an efficient combination of self-reflection, consistency across sampled responses, and probabilistic measures to comprehensively quantify the trustworthiness of a LLM response [13]. Unlike the other Evaluation models, TLM does not require a special prompt template. Instead, you simply use the same prompt you provided to your RAG LLM that generated the response being evaluated.

The prompt template used for TLM in our study looks like this:

```
Answer the QUESTION strictly based on the provided CONTEXT:
```

```
CONTEXT:
{context}

QUESTION:
{question}
```

## 3 Benchmark Methodology

To study the real-world hallucination detection performance of these Evaluation models/techniques, we apply them to RAG datasets from different domains. Each dataset is composed of entries containing: a user query, retrieved context that the LLM should rely on to answer the query, a LLM-generated response, and a binary annotation whether this response was actually correct or not.

Unlike other RAG benchmarks that study finer-grained concerns like retrieval-quality, faithfulness, or context-utilization, we focus on the most important overall concern in RAG: how effectively does each detection method **flag responses that turned out to be incorrect**. This is quantified in terms of *precision/recall* using the **Area under the Receiver Operating Characteristic curve (AUROC)**. A detector with high AUROC more consistently assigns lower scores to RAG responses that are incorrect than those which are correct. Note that none of our Evaluation models/techniques relies on the correctness-annotations, these are solely reserved for reporting the AUROC achieved by each scoring method.

We run all methods at default recommended settings from the corresponding libraries. Recall that both LLM-as-a-judge and TLM can be powered by any LLM model; our benchmarks run them with OpenAI's gpt-4o-mini LLM [14], a low latency/cost solution. If a scoring method failed to run for one or more examples in any dataset (due to software failure), we indicate this by a dotted line in the graph of results.

We report results over 6 datasets, each representing different challenges in RAG applications. Four datasets stem from the HaluBench benchmark suite [9] (we omitted the remaining datasets in this suite after discovering prevalent annotation errors within them). The other two datasets, **FinQA** [15] and **ELI5** [16], cover more complex settings.

## 4 Benchmark Results

### 4.1 FinQA

FinQA [15] is a dataset of complex questions from financial experts pertaining to public financial reports, where responses stem from OpenAI's GPT-4o LLM [17] in the version of this dataset considered in our study. The documents/questions in this dataset can be challenging even for humans to accurately answer, often requiring careful analysis of multiple pieces of information in the context and properly combining these pieces into a final answer (sometimes requiring basic arithmetic).

Here is an example from this dataset:

> *Query:*
> *What is the yearly interest incurred by the redeemed amount of junior subordinated debt, in thousands?*
>
> *Context:*
> *junior subordinated debt securities payable in accordance with the provisions of the junior subordinated debt securities which were issued on march 29 , 2004 , holdings elected to redeem the $ 329897 thousand of 6.2% ( 6.2 % ) junior subordinated debt securities outstanding on may 24 , 2013 . as a result of the early redemption , the company incurred pre-tax expense of $ 7282 thousand related to the immediate amortization of the remaining capitalized issuance costs on the trust preferred securities . interest expense incurred in connection with these junior subordinated debt securities is as follows for the periods indicated. Table:*
> ****************************************
> *( dollars in thousands ) | years ended december 31 , 2015 | years ended december 31 , 2014 | years ended december 31 , 2013 interest expense incurred | $ - | $ - | $ 8181*
> ****************************************
> *holdings considered the mechanisms and obligations relating to the trust preferred securities , taken together , constituted a full and unconditional guarantee by holdings of capital trust ii 2019s payment* *obligations with respect to their trust preferred securities . 10 . reinsurance and trust agreements certain subsidiaries of group have established trust agreements , which effectively use the company 2019s investments as collateral , as security for assumed losses payable to certain non-affiliated ceding companies . at december 31 , 2015 , the total amount on deposit in trust accounts was $ 454384 thousand .....*

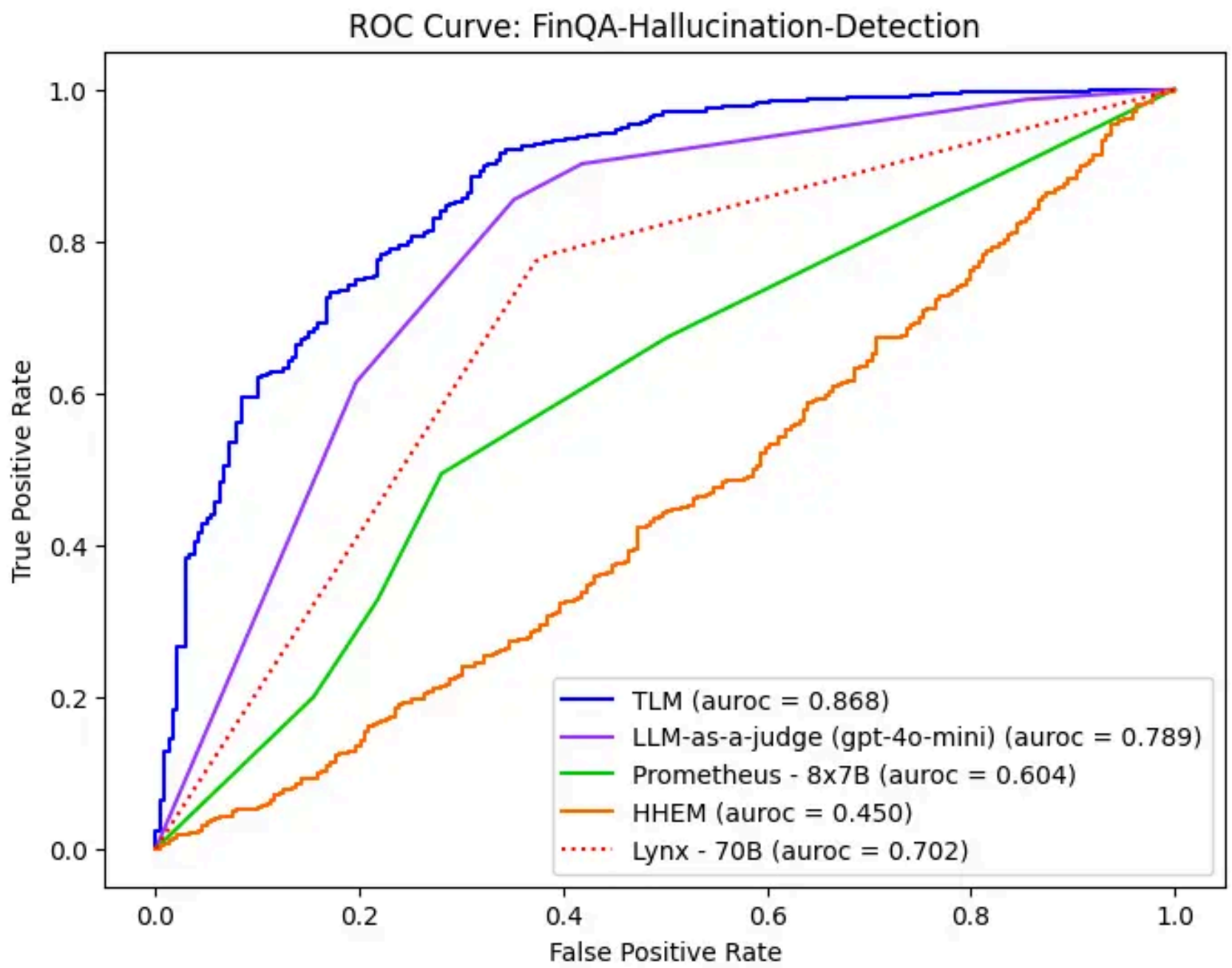

Figure 2. ROC curve (with AUROC values) for each Evaluation model on the FinQA dataset.

In this benchmark, almost all Evaluation models are reassuringly able to detect incorrect AI responses better than random chance (which would be AUROC = 0.5). Note that is far from guaranteed, given these methods must run in real-time without access to any annotations, ground-truth answers, or other extra information.

For FinQA, TLM and LLM-as-a-judge detect incorrect AI responses with the highest precision and recall. This may be due to the sophisticated nature of the questions/context, which custom trained models may not grasp as well as the OpenAI LLM we used to power TLM and LLM-as-a-judge in this benchmark.

## 4.2 ELI5

ELI5 [16] is a dataset that captures the challenge of breaking down complicated concepts into understandable/simplified explanations without sacrificing accuracy. Here we use a subset of ELI5 [18] dataset, where LLM responses were annotated as to their correctness. Responses that are annotated as incorrect exhibit issues such as: factual errors or misrepresentations, oversimplifying to the extent that inaccuracies are introduced, or making claims that are not supported by the retrieved context.

Here is an example from this dataset:

*Query:*
*What is the Theory of Nitrogen Fixation and how does it impact agriculture?*

*Context:*
*The Theory of Nitrogen Fixation explains the process by which certain organisms convert atmospheric nitrogen into a form that plants can use for growth. This process is primarily carried out by symbiotic bacteria, such as Rhizobium, which form nodules on the roots of leguminous plants. Nitrogen fixation is crucial for agriculture as it enriches the soil with nitrogen, reducing the need for chemical fertilizers and promoting sustainable farming practices. This natural process helps maintain soil fertility and*

*supports the growth of crops, leading to increased agricultural productivity.*

*Response:*
*The Theory of Nitrogen Fixation is like a magic trick where tiny helpers called bacteria turn air into food for plants. These bacteria live in little bumps on the roots of plants like corn and wheat. This magic helps plants grow better without needing a lot of extra plant food from factories, making farming kinder to the earth and helping farmers grow more food!*

*Annotation:*
*Bad Response*

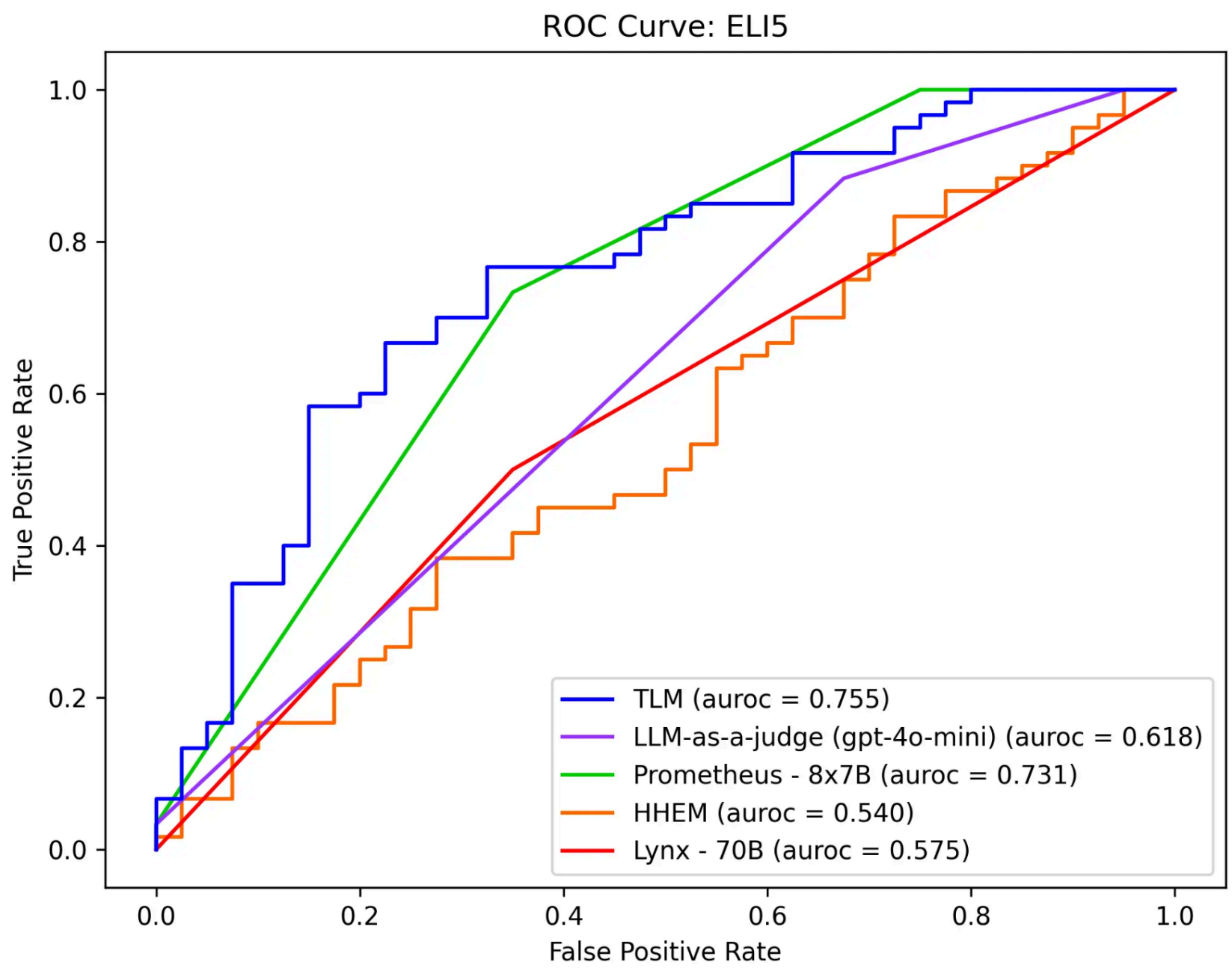

Figure 3. ROC curve (with AUROC values) for each Evaluation model on the ELI5 dataset.

In this benchmark, no method manages to detect incorrect AI responses with very high precision/recall, but Prometheus and TLM are more effective than the other detectors.

### 4.3 FinanceBench

FinanceBench [9] is a dataset reflecting the types of questions that financial analysts answer day-to-day, based on public filings from publicly traded companies including 10Ks, 10Qs, 8Ks, and Earnings Reports. Retrieved contexts contain financial documents of a company and the questions asked are clear-cut & straightforward to answer by reading relevant context (unlike FinQA where answers often require non-trivial reasoning).

Here is an example from this dataset:

*Query:*
*What is the FY2021 operating cash flow ratio for CVS Health? Operating cash flow ratio is defined as: cash from operations / total current liabilities.*

*Context:*
*Index to Consolidated Financial Statements*
*ConsolidatedBalanceSheets*
*AtDecember31,*
*In millions, except per share amounts*
*2021*
*2020*
*Assets:*

*Cash and cash equivalents*
*$ 9,408*
*$ 7,854*
*Investments*
*3,117*
*3,000*
*Accounts receivable, net*
*24,431*
*21,742 ...*

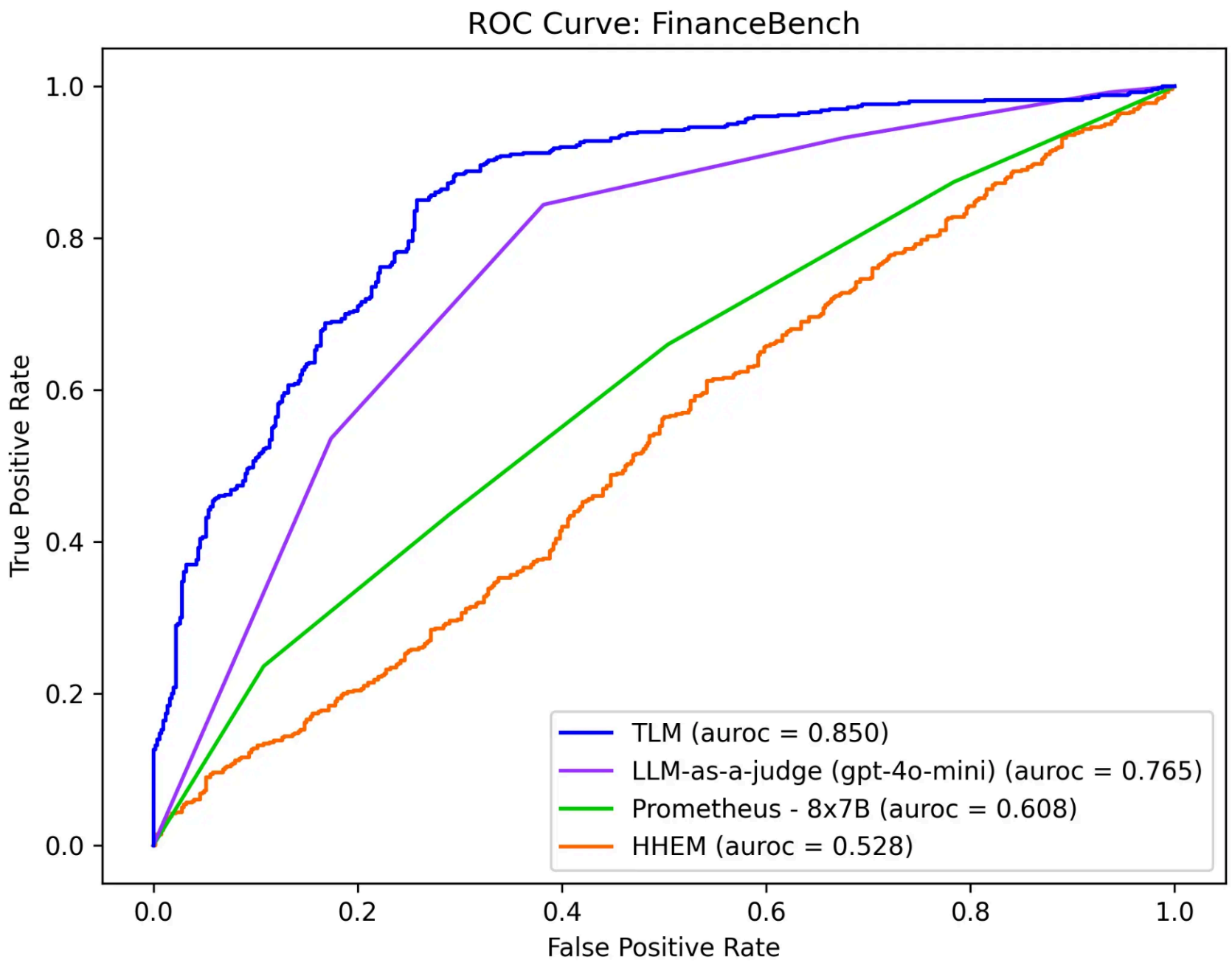

Figure 4. ROC curve (with AUROC values) for each Evaluation model on the FinanceBench dataset.

For FinanceBench, TLM and LLM-as-a-judge detect incorrect AI responses with the highest precision and recall. This matches the previous findings for FinQA, even though the two datasets contain different types of information and user queries.

### 4.4 PubmedQA

PubmedQA [9] is a dataset where context comes from PubMed (medical research publication) abstracts that are used by the LLM to answer biomedical questions. Here is an example query from this dataset that the LLM must answer based on relevant context from a medical publication:
*A patient with myelomeningocele: is untethering necessary prior to scoliosis correction?*

In this benchmark, Prometheus and TLM detect incorrect AI responses with the highest precision and recall. For this application, these Evaluation models are able to catch hallucinations very effectively.

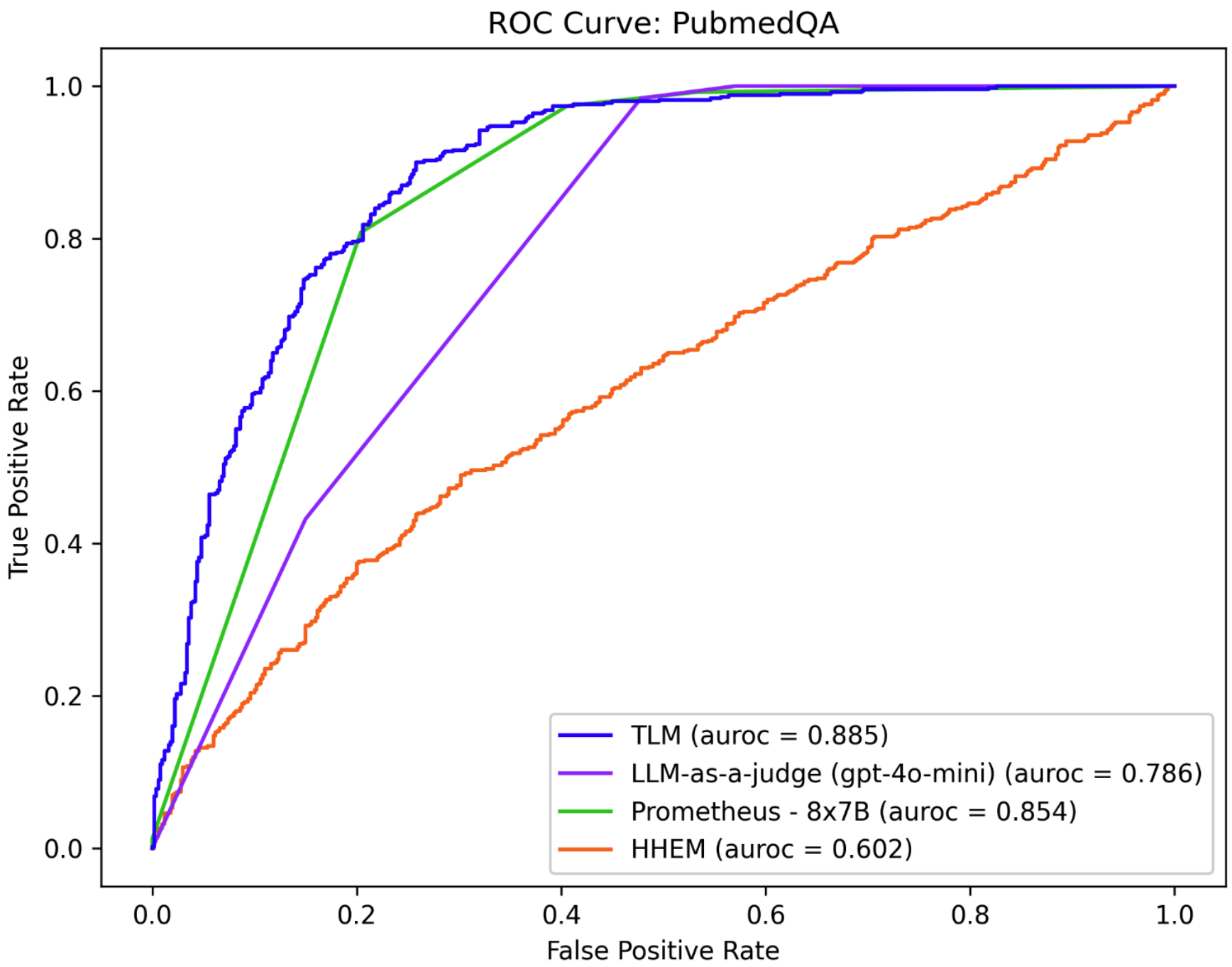

Figure 5. ROC curve (with AUROC values) for each Evaluation model on the PubmedQA dataset.

## 4.5 CovidQA

CovidQA [9] is a dataset to help experts answer questions related to the Covid-19 pandemic based on the medical research literature. The dataset contains scientific articles as retrieved context and queries such as: *What is the size of the PEDV genome?*

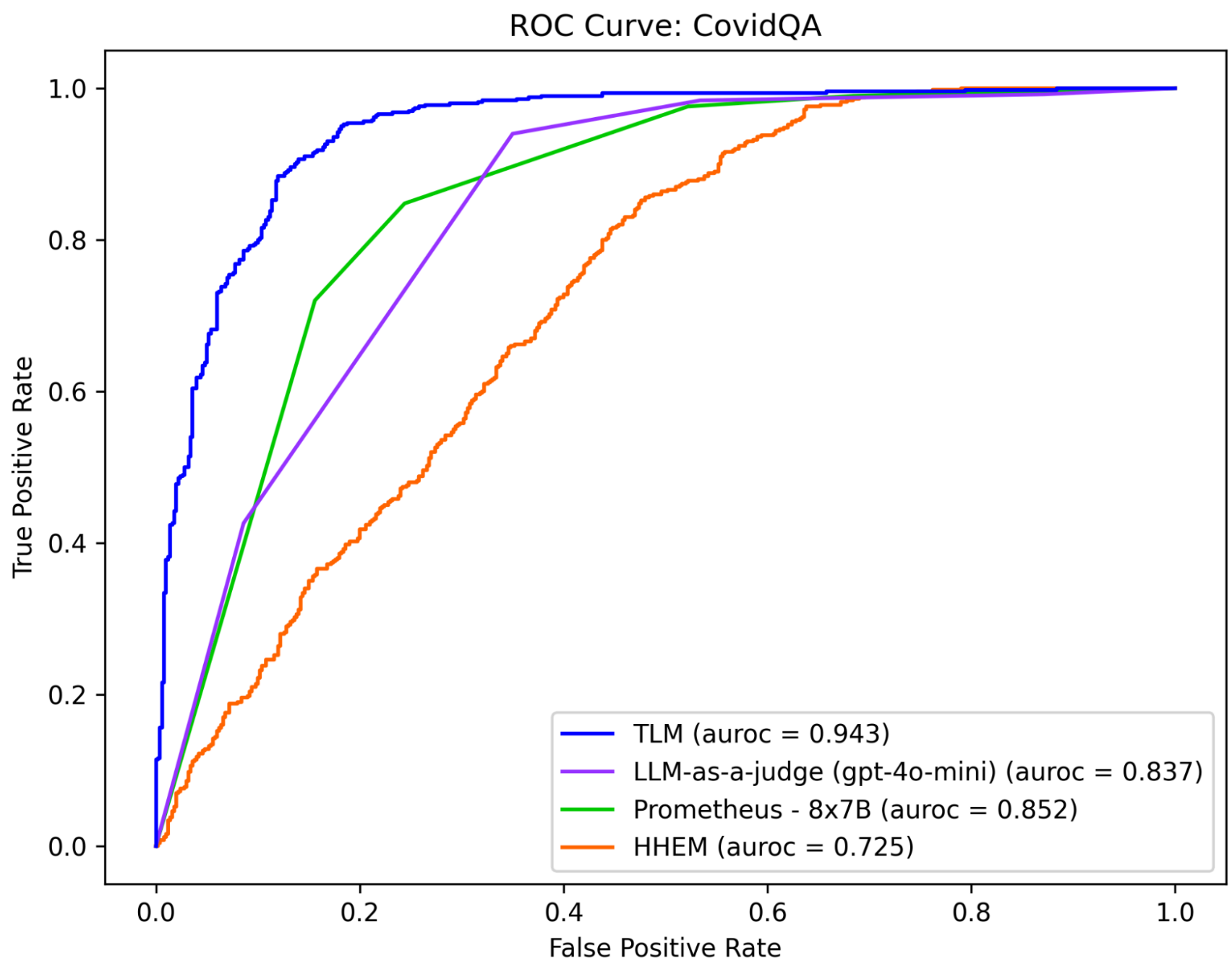

Figure 6. ROC curve (with AUROC values) for each Evaluation model on the CovidQA dataset.

In this benchmark, TLM detects incorrect AI responses with the highest precision and recall, followed by Prometheus and LLM-as-a-judge. For this application, today's top Evaluation models provide a reliable way to catch hallucinations.

### 4.6 DROP

Discrete reasoning over paragraphs (DROP) [9] consists of passages retrieved from Wikipedia articles and questions that require discrete operations (counting, sorting, etc.) and mathematical reasoning to answer.

For example, given a passage about the game between the Cowboys and Jaguars, a question from the dataset is:

> *Which team scored the least points in the first half?*

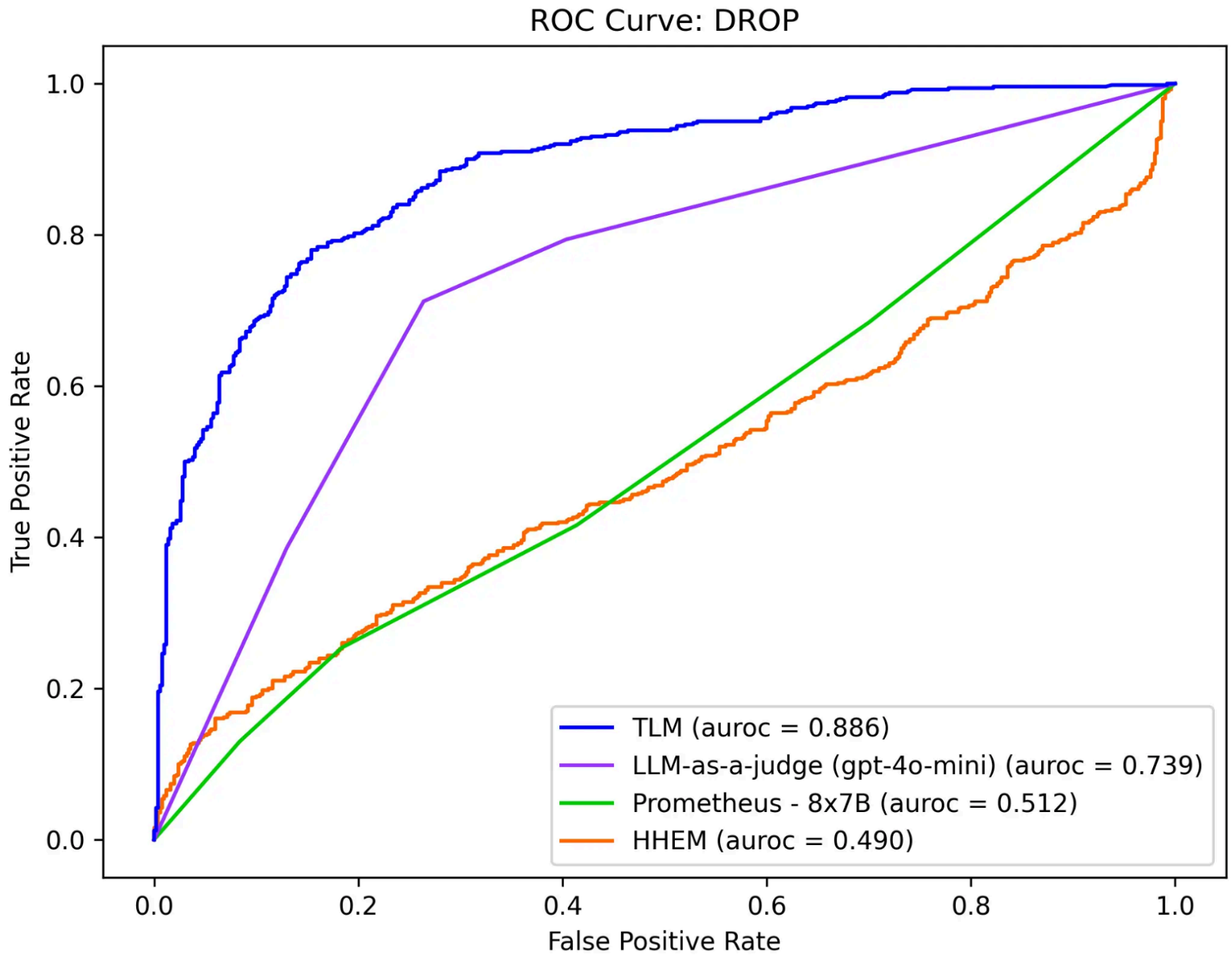

Figure 7. ROC curve (with AUROC values) for each Evaluation model on the DROP dataset.

In this benchmark, TLM detects incorrect AI responses with the highest precision and recall, followed by LLM-as-a-judge. No other Evaluation model appears very useful in this more challenging application.

## 5 Conclusion

Our study presented one of the first benchmarks of real-time Evaluation models in RAG. We observed that most of the Evaluation models could detect incorrect RAG responses significantly better than random chance on some of the datasets, but certain Evaluation models didn't fare much better than random chance on other datasets. Thus carefully consider your domain when choosing an Evaluation model.

Some of the Evaluation models in this study are specially trained custom models. Since these custom models were trained on errors made by certain LLMs, their future performance remains unclear when future LLMs make different types of errors. In contrast, evaluation techniques like LLM-as-a-Judge or TLM can be powered by *any* LLM and should remain relevant for future LLMs. These techniques require no data preparation / labeling, nor infrastructure to serve a custom model.

Besides the Evaluation models covered in this study, other techniques have been used for detecting hallucinations. One previous study [19] benchmarked alternative techniques, including **DeepEval**, **G-Eval**, and **RAGAS**, finding that Evaluation models like TLM detect incorrect RAG responses with universally higher precision/recall.